\documentclass[fleqn]{optica-article}

\journal{opticajournal} 

\usepackage[utf8,utf8x]{inputenc}
\usepackage{graphicx}
\usepackage[table]{xcolor}
\usepackage{tikz}
\usepackage{appendix}
\usepackage[colorinlistoftodos]{todonotes}
\usepackage[english]{babel}
\usepackage[utf8x]{inputenc}
\usepackage{xcolor}

\usepackage{listings}

\usepackage{amsmath}

\usepackage{amsthm}

\usepackage{bbding}

\usepackage{changes}
\usepackage{hyperref}
\usepackage{multirow}
\usepackage{array}

\articletype{Research Article}

\newtheorem{example}{Example}

\begin{document}
\pagestyle{plain}

\title{Handling Wikidata Qualifiers in Reasoning}

\author{Sahar Aljalbout,\authormark{1 *} Gilles Falquet,\authormark{2} and Didier Buchs\authormark{3}}

\address{\authormark{1, 2,3 } Centre Universitaire d'informatique, University of Geneva, Switzerland\\
}

\email{\authormark{*}sahar.aljalbout@unige.ch} 


\begin{abstract*}
Wikidata is a knowledge graph increasingly adopted by many communities for diverse applications. Wikidata statements are annotated with qualifier-value pairs that are used to depict information, such as the validity context of the statement, its causality, provenances, etc. Handling the qualifiers in reasoning is a challenging problem. When defining inference rules (in particular, rules on ontological properties (x subclass of y, z instance of x, etc.)), one must consider the qualifiers, as most of them participate in the semantics of the statements. This poses a complex problem because a) there is a massive number of qualifiers, and b) the qualifiers of the inferred statement are often a combination of the qualifiers in the rule condition. In this work, we propose to address this problem by a) defining a categorization of the qualifiers b) formalizing the Wikidata model with a many-sorted logical language; the sorts of this language are the qualifier categories. We couple this logic with an algebraic specification that provides a means for effectively handling qualifiers in inference rules. Using Wikidata ontological properties, we show how to use the MSL and specification to reason on qualifiers. Finally, we discuss the methodology for practically implementing the work and present a prototype implementation. The work can be naturally extended, thanks to the extensibility of the many-sorted algebraic specification, to cover more qualifiers in the specification, such as uncertain time, recurring events, geographic locations, and others. 
\end{abstract*}




\section{Introduction} 
\newcommand\SP{\textsf{st}}
The advancement of artificial intelligence has sparked interest in graph-based knowledge representation, especially knowledge graphs (KG). They are essential components of intelligent assistants such as Apple's Siri and Amazon's Alexa \cite{haase2017alexa}, of the question-answering systems of modern search engines such as Google and Microsoft's Bing, and of the new expert systems in the style of IBM's Watson \cite{ferrucci2010building}.

Wikidata is one of the most successful publicly available knowledge graphs. The main representation object of the Wikidata model is the \emph{statement}. A statement\footnote{https://www.wikidata.org/wiki/Wikidata:Glossary} claims that an entity (the subject) has a certain property with a given value. In addition, a statement can be qualified with qualifier-value pairs and should have at least one reference (information about sources). For instance, the fact that \textsf{George C. Scott}  was married to \textsf{Colleen Dewhurst} from \textit{1960} until their divorce in \textit{1965} is represented in Wikidata by a statement that we will denote as:
\begin{equation}
\label{eq:scott}
\begin{array}{l}
 (\textsf{George C. Scott, spouse, Colleen Dewhurst}) \\
\quad [\textsf{start time}: \textit{1960},  \textsf{end time}: \textit{1965},  \textsf{end cause}: \textit{divorce}] \\ 
\end{array}
\end{equation}

where \textsf{George C. Scott}, \textsf{spouse},  and \textsf{Colleen Dewhurst} are the subject, property, and value, respectively, and  \textsf{start time}, \textsf{end time}, \textsf{end cause} are qualifiers with value \textit{1960}, \textit{1965}, and \textsf{divorce}, respectively \footnote{In fact, each entity and property is represented by an IRI in Wikidata, but for the sake of readability, we will use their English labels instead.}. The statement is also supported by one reference\footnote{Most Wikidata statements are supported by at least one reference that we omit in the examples to keep them compact.}.

The Wikidata statements form a multi-qualified\footnote{also known as multi-attributed knowledge graph} knowledge graph because each statement may have several values for each qualifier.

\subsection{Reasoning without qualifiers in Wikidata}Although Wikidata already contains a rich information set, much more could be derived by applying inference rules to the existing statements. In particular, Wikidata has ``ontological''\footnote{they are not axiomatized} properties that can directly lead to the definition of inference rules. Ontological properties, such as \textsf{instance of, subclass of, subproperty of, equivalent property, etc.}, correspond to the usual modeling primitives found in conceptual or ontological modeling languages, such as the description logics or RDFS. For some of these languages, e.g., RDFS or OWL2 RL, a set of inference rules can produce all the formulas entailed by a set of axioms. A large part of these rule sets can be adapted to provide ontological inference rules for Wikidata. For instance, the RDFS subClassOf rule \textsf{rdfs9}\footnote{\url{https://www.w3.org/TR/rdf11-mt/}}
\begin{quote}
\begin{tabular}{|c|c|c|}
\hline
           & IF the graph $S$ contains & THEN $S$ RDFS entails \\
\hline
     \texttt{rdfs9} &  \texttt{xxx rdfs:subClassOf yyy} &  \texttt{zzz rdf:type yyy }\\
           &  \texttt{zzz rdf:type xxx}  & \\
\hline
\end{tabular}\\
\end{quote}
would become
\begin{itemize}
    \item[] IF ($x$, \textsf{subclass of}, $y$) and ($z$, \textsf{instance of}, $x$)  THEN ($z$,  \textsf{instance of}, $y$)
\end{itemize}
in Wikidata.

Wikidata also defines several types of property constraints\footnote{\url{https://www.wikidata.org/wiki/Wikidata:WikiProject_property_constraints}}  \textsf{(subject type, value type, symmetry, etc.)} that are primarily intended to check the integrity of the graph. These constraints could be used to generate inference rules that add statements to satisfy the constraints. 
For instance, given a symmetry constraint on property $P$ and a statement $(x, P, y)$, if the symmetric statement $(y, P, x)$ is not in the graph, one can either raise an error condition (the constraint checking approach) or infer $(y, P, x)$ to satisfy the constraint (the inference approach).

In addition to these ontological rules, multiple potential rules originate from the different knowledge domains covered by Wikidata. Moreover, any application using Wikidata may also need specific inference rules.

\subsection{Reasoning with qualifiers in Wikidata}
While it is relatively straightforward to design inference rules on
the subject, property, and value of a Wikidata statement, as shown above, things become more complicated when one tries to take the qualifiers into account because there is no unique and clearly established way to do so\footnote{As mentioned, for instance, in \url{https://www.wikidata.org/wiki/Wikidata:WikiProject_Reasoning}}. The following examples show a variety of situations that require different treatments of the qualifiers. 

The \textsf{spouse} property is symmetric. Therefore, from statement (\ref{eq:scott}), that represents the spousal relationship of  \textsf{George C. Scott}, we can infer:
\begin{equation}
\label{eq:dewhurstscott}
\begin{array}{l}
 (\textsf{Colleen Dewhurst, spouse, George C. Scott }) \\
\quad [\textsf{start time}: \textit{1960},  \textsf{end time}: \textit{1965},  \textsf{end cause}: \textit{divorce}] \\ 
\end{array}
\end{equation}

Here, the qualifiers and their values can be directly copied into the inferred statement. But this is not always the case. For instance, assuming that the \textsf{part of} property is transitive, from the statements

\begin{equation}
\label{eq:northc}
\begin{array}{l}
(\textsf{Province of North Carolina, part of, Southern Colonies })\\
\quad[\textsf{start time}: \textit{10 May 1775}, \textsf{end time}: \textit{4 July 1776}] \\ 
\end{array}
\end{equation}
and
\begin{equation}
\label{eq:southco}
\begin{array}{l}
 (\textsf{Southern Colonies, part of, British Empire }) \\
\quad[\textsf{start time}: \textit{9 June 1732}, \textsf{end time}: \textit{4 July 1776}
] \\ 
\end{array}
\end{equation}
one can infer
\begin{equation}
\label{eq:nothco2}
\begin{array}{l}
 (\textsf{Province of North Carolina, part of, British Empire })\\
\quad[\textsf{start time}: \textit{10 May 1775},   \textsf{end time}: \textit{4 July 1776}] \\ 
\end{array}
\end{equation}
In this case, the validity time of the inferred statements, defined by \textsf{start time} and \textsf{end time}, is the intersection of statement (\ref{eq:northc}) and (\ref{eq:southco}) validity times.

 \begin{figure}[htbp]
    \centering
    \fbox{\includegraphics[width=14cm, ]{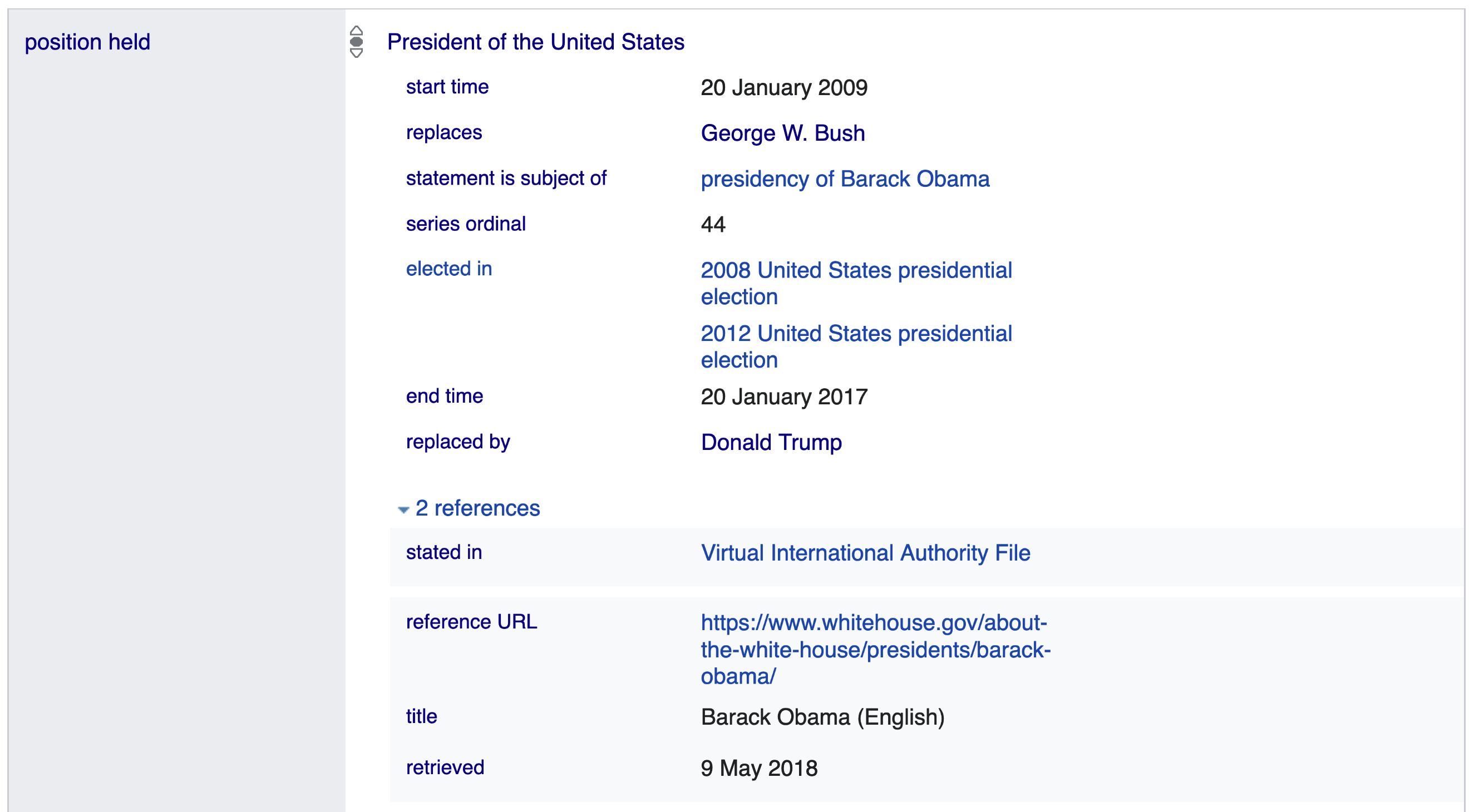}}
    \caption{A statement with the \textbf{position held} property and qualifiers for the item Q76 (\textbf{Barack Obama})}
    \label{fig:barack}
\end{figure} 
Furthermore, we find statements annotated with what we propose to call sequence qualifiers such as \textsf{replaces} and \textsf{replaced by}, for example the \textsf{position held} property. 
The example of the Barack Obama presidency in Fig \ref{fig:barack}  illustrates well the use of these qualifiers. Using the \textsf{replaces} qualifier in Fig \ref{fig:barack}, we could infer a statement such as (6):
\begin{equation}
\begin{array}{l}
 (\textsf{George W. Bush, position held, President of the United States})[ \\
\quad\textsf{end time}: \textit{20 January 2009}, 
\quad\textsf{replaced by}: \textit{Barack Obama} 
] \\ 
\end{array}
\end{equation}
In the United States, the presidency of a particular president starts the day the presidency of the previous one ends. Based on the \textsf{start time} value of Fig \ref{fig:barack}, we can infer the \textsf{end time} in (6). In other countries, the beginning of a new president's presidency might take longer. In such cases, using a theory of uncertain time would be more adequate. Wikidata contains qualifiers for such a theory: \textsf{earliest time} and \textsf{latest time}. As for the sequence qualifiers: using the \textsf{replaces} qualifier of Fig \ref{fig:barack}  to infer (6) implies that (6) should contain a \textsf{replaced by} qualifier.

This description of operations needed for different qualifiers shows that we need a well-formalized theory for reasoning on qualifiers, notably because Wikidata defines more than 9000 qualifiers, of which 200 appear in more than 10000 statements 
\footnote{These figures were obtained by counting the number of lines in qualifier files (in RDF/Turtle) extracted from a Wikidata RDF dump of 2022-02-16 and available at \url{http://ke.unige.ch/wikidata/WikidataDump/pq/}}. 
We formalize this theory in the algebraic specification in Section 3.

\subsection{Proposed Approach} The above examples show that taking qualifiers into account in the reasoning poses (at least) two problems.

\begin{itemize}
    \item[\textbf{P1:}] Handling qualifiers in inference rules highly depends on the properties and the qualifiers participating in a rule. Due to the massive number of qualifiers, specifying the behavior of each qualifier in each rule is out of the question. Instead, we need a more abstract view of the qualifiers.    
    \item[\textbf{P2:}] Some qualifiers need complex computations, such as computing the intersection of certain or uncertain time intervals, 
    calculating geographic locations, determining causalities, tracing provenances, etc.
    Therefore, it is necessary to provide a uniform and expressive handling of the different types of qualifiers by defining formal theories. 
\end{itemize} 

To have an abstract view of the qualifiers (P1), we propose a categorization of the Wikidata qualifiers. Then, we present a many-sorted logic (MSL) in which a sort represents each qualifier category. For instance, qualifiers such as \textsf{start time, end time} participate in the specification of the temporal validity context sort of a statement; \textsf{has cause, end cause} participate in the causality sort; etc. We show that it is possible to classify the most frequently used qualifiers in a limited number of sorts and refer to them instead of the individual qualifiers in the expression of inference rules (section 2).

To resolve (P2), we couple the logic with an algebraic specification: functions and axioms in a specific module of an algebraic specification will define the test and combination operations specific to each qualifier category.

The paper aims to show the methodology to handle qualifiers in reasoning using MSL and algebraic specification. Moreover, the approach is extensible because we can apply it to other categories of qualifiers.

\subsection{Structure of this Paper}
Section 2 exemplifies the results of a study of Wikidata qualifiers. We show that a few categories are sufficient to group the most frequently used qualifiers. Section 3 describes the approach used to represent a multi-attributed graph as a set of MSL formulas (in an MS theory) and an algebraic specification. We show how to apply it to Wikidata. In addition,  we explain the design of the algebraic specification. Using Wikidata ontological properties, we show in Section 4 how to exploit the MSL and specification to reason on qualifiers. Section 5  briefly explains the techniques to implement the approach and perform inferences on a Wikidata dump. Section 6 discusses the extensibility of the work. Finally, Sections 7 and 8 discuss related works and conclude the paper.

\section{Categorizing Wikidata qualifiers}
Wikidata uses qualifiers to describe further or refine a given statement. Fig \ref{fig:barack} shows a statement with $\mathsf{position \ held}$ property. The qualifiers are used for: 
\begin{enumerate}
    \item  Describing the validity time of the statement: from \textit{20 January 2009} until \textit{20 January 2017}.
    \item Specifying the ordering of the presidency: Barak Obama is the 44$^\text{th}$ president of the USA ($\mathsf{series\ ordinal}$), he succeeds George Bush (\textsf{replaces}) and is followed by Donald Trumph (\textsf{replaced by}).
    \item Stating the references of the statement: this statement is supported by two references specified by their source document (\textsf{stated in} and \textsf{reference URL}); title (\textsf{title}); and retrieval date (\textsf{retrieved}). 
    \item Providing more details about the statement (\textsf{elected in}, \textsf{statement is subject of}). 
\end{enumerate}

These qualifiers vary from one statement to another. A study of the most prominent Wikidata qualifiers\footnote{We used a dump retrieved on 16/02/2022}revealed that a large part of them serve to:
\begin{enumerate}
    \item Specify some temporal or spatial or other restriction on the \emph{validity context} of the statement (\textsf{valid in}, \textsf{start time}, \textsf{end time}, etc.) or
    \item Annotate the statement with additional information that qualifies, among others, the statement's  cause or provenance, the position of the subject or object in a sequence, etc. 
\end{enumerate}
In his paper \cite{Patel-Schneider2018}, the author distinguishes between additive and contextual qualifiers in Wikidata. Additive qualifiers add extra information about the fact (i.e., Wikidata statements), such as \textsf{replaces} and \textsf{replaced by} in figure 1. Contextual qualifiers contextualize the underlying fact, i.e., they limit the contexts in which the underlying fact is true, such as \textsf{start time} and \textsf{end time}. In this paper, we refine this description. First, we exemplify that the contextual qualifiers (aka. validity contexts qualifiers) are various in Wikidata: they are not limited to time and space but can also be domain related. Then, we show that additional qualifiers can be further classified into causalities, sequence, annotations, and provenance and some of them can be effectively handled in reasoning.

After exploring the qualifiers of Wikidata\footnote{The file at \url{http://ke.unige.ch/wikidata/Statistics/qualifier-prominence.csv} contains the prominence values of the qualifiers and the categorization of the most frequent ones},  we identified the five categories described in the following subsections 
\begin{table} [hbt] 
\begin{tabular}{ |p{2.8cm}||p{6cm}| r |  }
 \hline
 Qualifier category & Qualifier name  & Prominence\\
 \hline
 Validity Context  & \textsf{point in time} (P585) & 9\,883\,140 \\
 & \textsf{start time} (P580) & 6\,853\,221 \\
   & \textsf{end time} (P582) & 3\,573\,088  \\
 & \textsf{applies to jurisdiction} (P1001) & 1\,231\,796  \\
& \textsf{applies to part} (P518) & 840\,468 \\
& \textsf{valid in period } (P1264)& 781\,810   \\
& \textsf{latest date}  (P1326) & 283\,290   \\
& \textsf{earliest date} (P1319)& 241\,435  \\

  \hline
 Causality   & \textsf{end cause} (P1534) & 135\,146 \\
 & \textsf{has cause} (P828) & 26\,177 \\

 \hline
 Sequence & \textsf{series ordinal} (P1545) & 157\, 611 \,677 \\
 & \textsf{follows} (P155) & 899\,272 \\
 & \textsf{followed by} (P156) & 898\,373 \\
  & \textsf{replaces} (P1365) & 200\,523 \\
  & \textsf{replaced by} (P1366) & 177\,909 \\
\hline
 Provenance  
 & \textsf{object named as} (P1932) & 9\,962\,214 \\
 & \textsf{determination method} (P459) & 7\,909\,851 \\
 & \textsf{subject named as} (P1810) & 2\,436\,681 \\
 & \textsf{criterion used} (P1013) & 1\,280\,779 \\
 & \textsf{sourcing circumstances} (P1480) & 228\,090 \\
 \hline
\end{tabular}
\label{Tab:1-qualifiers categories}
\caption{\label{Tab:qualifiers categories} Prominence of the some qualifiers of each category
(some properties have a very high prominence but are domain-dependant. We don't present them in this table).} 

\end{table}

\subsection{Validity contexts}
The validity context qualifiers affect the semantics of the statement by restricting its truth to a specific context. For instance, when we say that \textsf{Barack Obama was the president of the United States from 20 January 2009 till 20 January 2017}, we restrict the validity of his presidency to the temporal interval [20/1/2009, 20/1/2017]. 

Validity contexts are not restricted to time; we also have validity space. For example, according to Wikidata, \textsf{Beatrice Saubin was convicted of drug trafficking in Malaysia}\footnote{https://www.wikidata.org/wiki/Q42304190}. Hence, the validity of her conviction is restricted to Malaysia, which makes Malaysia a validity space. The \textsf{applies to jurisdiction} qualifier typically defines a territorial entity in which the statement is true, hence a spatial validity context. 
Other qualifiers such as \textsf{applies to work} or \textsf{applies to taxon} are also context indicators.
Therefore the validity context of a statement 
can be constituted of several dimensions, such as time, space, work (e.g., for statements about characters in a book), taxon (e.g., for statements about the toxicity of a chemical product for some organism), etc. 

Validity context qualifiers are among the most frequently used. Table \ref{Tab:qualifiers categories} shows some of the most frequently used validity qualifiers\footnote{The table does not show the qualifiers that are very specific, such as \textsf{astronomical filter}, \textsf{chromosome}, \textsf{found in taxon}, etc. Although very frequent, they qualify only a limited number of properties that belong to a specific domain (astronomy, genetics, biology).  }.

\subsection{Causality}
Wikidata contains causality qualifiers used to assert the causality of some statements' beginning/end, such as a marriage ending, or a political or a social event. In example (1) of the spousal relationship (i.e., between George C. Scott and Colleen Dewhurst), the statement is qualified with an \textsf{end cause} qualifier. It is used to express that this marriage ended due to divorce. \textit{Divorce} is the value of the \textsf{end cause} qualifier. Wikidata uses mostly two causality qualifiers \textsf{has cause} and \textsf{end cause}.  Table \ref{Tab:qualifiers categories} shows the causality qualifiers: for each qualifier, its ID, and its prominence. The \textsf{end cause} qualifier is used more frequently than the \textsf{has cause} qualifier.
\subsection{Sequence}
A typical example of sequence qualifiers is the Wikidata page of a politician. Figure \ref{fig:barack} shows the Wikidata page of Barack Obama and especially the \textsf{position held} property. Among the qualifiers used, we can spot two sequence qualifiers: \textsf{replaces} and \textsf{replaced by}. The latter is used to name the president who was elected after Barack. The former is used to name the president before him. Sequence qualifiers behave like pointers pointing to the element that comes before the subject and the element after the subject. Table \ref{Tab:qualifiers categories} shows the sequence qualifiers. For each qualifier, its ID and its prominence are shown. We observe that the \textsf{follows/followed} by qualifiers are used more then \textsf{replaces/replaced by}.

\subsection{Provenance}
 Wikidata uses some qualifiers that provide information about the information derivation process: determination method, criterion used, sourcing circumstances, etc. We group these qualifiers under the category of provenance.

\subsection{Annotations}
We group all other qualifiers into one category called annotations. They provide additional information to the statements. Annotations in Wikidata can also be divided into subcategories:
\begin{itemize}
    \item Constraint annotations: they are usually used to qualify a statement with a \textsf{property constraint} (P2302). Examples of these properties: \textsf{exception to constraint}  (P2303), \textsf{constraint status} (P2316), \textsf{separator} (P4155), etc.
    \item Others: attributes used to add more information about a particular topic such as: \textsf{subject named as, object has role, together with etc.} 
     \end{itemize}


\section{A many-sorted logic for Wikidata}
\theoremstyle{definition}
\newtheorem{definition}{Definition}
\newtheorem*{remark}{Remark}
In this section, we propose to use a many-sorted logic (MSL) approach to formalize the semantics of Wikidata knowledge graph. To this end we will map every Wikidata statement to an MSL atom and we will add axioms, in the form of an algebraic specification, to specify the behavior of the qualifiers in reasoning.  Let us start by defining a MSL representation technique for multi-qualified knowledge graphs.
\paragraph{Preliminaries} A many-sorted logic is a logic, in a first-order language, in which the universe is divided into subsets called sorts. For example, one might divide the universe of discourse into different kinds (called sorts) of animals and plants \cite{walther1985mechanical}, \cite{cohn1985solution}. In a many-sorted logic language the arity of a predicate is a sequence of sorts $s_1 \times \cdots \times s_k$ (indicating that the i$^{th}$ argument must be of sort $s_i$). Similarly, a function arity is a sequence  $s_1 \times \cdots \times s_l \to s$ ($s$ is the sort of the result)

\subsection{MSL model for multi-qualified knowledge graphs}\label{subsec:mslmodel}
\begin{definition}

We define a multi-qualified knowledge graph as a tuple $(V, E, P, D, S)$ where 
\begin{itemize}
\item $V$ is the set of values (nodes) of the graph, 
\item $E \subseteq V$ is a set of IRIs denoting real world entities, 
\item $P \subseteq E$ is a set of properties denoting binary relations, 
\item $D \subseteq V$ is a set of data value denotations , 
\item $V= E \cup D$ and $D \cap E = \emptyset$,
\item $S$ is a set of statements of the form
$$(s, p, v, \{q_1: v_1, ..., q_n: v_n\})$$
where $s \in E$ is the statement's subject, $p \in P$ is the property, $v \in V$ is the value, $q_i:v_i \in P \times V$ $(i=1,n)$ are qualifier-value pairs. 
\end{itemize}
\end{definition}

As  shown in section 2, the qualifiers of a multi-qualified knowledge graph such as Wikidata can be assigned to different categories. In order, to handle effectively all these qualifiers in reasoning, we propose to consider each qualifier category $C$ (for instance, validity context, causality, sequence) as a sort $s_C$ in a many-sorted logic language and to associate an object of sort $s_C$ to each statement. This object will represent the qualifiers belonging to the category $C$. For instance, in Wikidata, the causality is described through the qualifiers \textsf{has cause} and \textsf{end cause}, hence a value of the $s_\mathsf{causality}$ sort will be constructed for each statement by taking the values of the \textsf{has cause} and \textsf{end cause} qualifiers (if they exist). 

\begin{definition}
A MSL vocabulary $\Sigma$ for a multi-qualified knowledge graph $G = (V, E, P, D, S)$ is comprised of
\begin{itemize}
    \item a set of sorts that contains the sorts \textsf{value}, \textsf{entity}, \textsf{property}, \textsf{datavalue}, and sorts  $\mathsf{s}_1, \ldots, \mathsf{s}_n$ for representing the desired qualifier categories 
    \item a subsort relation that contains $\mathsf{entity} \le \mathsf{value}$, $\mathsf{property} \le \mathsf{entity}$,
    \item a set of constants of sort \textsf{value} that includes all the IRIs in $V$
    \item a set of constants of sort \textsf{entity} that includes all the IRIs in $E$
    \item  a set of constants of sort \textsf{property} that includes all the  IRIs in $P$
    \item  a set of constants of sort \textsf{datavalue} that includes all the  elements of $D$
    \item a predicate symbol $st$ with arity $\mathsf{entity} \times \mathsf{property} \times \mathsf{value} \times \mathsf{s}_1 \cdots \times \mathsf{s}_n$
    \item a set of function symbols together with their arities
\end{itemize}
\end{definition}
\begin{definition}
    Given a multi-qualified knowledge graph $G= (V, E, P, D, S)$ and a vocabulary $\Sigma$ for $G$, a many-sorted representation of a statement $(s,p,v,\{q_{1}: v_{1}, \ldots , q_{n}: v_{n} \}$ of $S$ with $\Sigma$ is a term 
    $$st(s, p, v, t_1, \ldots, t_k)$$
     where each $t_i$ is a ground term over $\Sigma$ of sort $s_i$
\end{definition}
The $t_i$'s terms correspond to the construction of the sort values from the qualifier values.

\begin{definition} 
    A many-sorted representation of a knowledge graph $G$ is made of
    \begin{enumerate}
        \item A many-sorted vocabulary  $\Sigma$ for $G$
        \item A many-sorted representation of each graph statement with $\Sigma$
        \item A first order theory Spec over $\Sigma$
    \end{enumerate}
\end{definition}

The role of the $\mathsf{Spec}$ theory is to specify the qualifier categories theories through axioms on the functions that act on the corresponding sorts. In particular it must define the semantics of
\begin{itemize}
    \item The functions that construct sort values from qualifiers values (of sort $\mathsf{value}$). 
    \item The functions that will be used in inference rules to access and combine the sorts values.
\end{itemize}

Since the interpretation domain we are considering is a many-sorted set, it is natural to specify the functions with the algebraic specification technique. Indeed, the models of algebraic specifications are precisely $\Sigma$-algebras (composed of a many-sorted set and many-sorted functions)\cite{sannella_foundations_2012}. Moreover, algebraic specifications can be organised into modules. In our case, we define a module for each qualifier category. Each module provides the theory chosen for the corresponding qualifier category in this setting. Moreover , as the framework allow this, they can be extended by adding another axioms if necessary.

\subsection{MSL model applied on Wikidata}


The Wikidata qualifiers analysis (Section 2)  shows that they are used to express five categories: validity, causality, sequence, annotations, and provenance. Therefore an MSL vocabulary for the representation of a Wikidata graph will comprise five sorts $\mathsf{s_{validity}}$, $\mathsf{s_{causality}}$, $\mathsf{s_{sequence}}$, $\mathsf{s_{annotations}}$, $\mathsf{s_{provenance}}$, in addition to the \textsf{value}, \textsf{entity}, \textsf{property}, and \textsf{datavalue} sorts.

Thus the arity of the \textsf{st} predicate that represents a Wikidata statement becomes  
$$\mathsf{entity} \times \mathsf{property} \times \mathsf{value} \times \mathsf{s_{validity}} \times \mathsf{s_{causality}} \times \mathsf{s_{sequence}} \times \mathsf{s_{annotations}} \times \mathsf{s_{provenance}}$$

%

\begin{example}\label{ex:scott}
To represent  statement (1) of the Introduction about George C. Scott's spouse we need functions to build its validity context and its causality. Since the validity is temporal in this case, it is convenient to introduce two additional sorts $\mathsf{s_{timeInterval}}$  and  $\mathsf{s_{instantTime}}$  to represent the temporal dimension of the validity. The statement could then be represented by the atom
$$
 \begin{array}{l}
 \mathsf{st}(\textsf{George C.Scott},  \mathsf{spouse}, \textsf{Colleen dewhurst}, \\
 \qquad \mathsf{timeValidity}( \mathsf{interval}( \mathsf{instant(25/11/1991)}, \mathsf{instant(11/5/2001)})), \\
 \qquad \mathsf{addEndCause}(\mathsf{\{divorce\}}, \varepsilon_C),\\
 \qquad \varepsilon_\mathsf{S}, \varepsilon_\mathsf{P}, \varepsilon_\mathsf{A})
 \end{array}
 $$ 
 where the functions have arities
\begin{itemize}
    \item $\mathsf{timeValidity:s_{timeInterval} \to s_{validity}}$
    \item $\mathsf{interval : s_{instantTime} \times s_{instantTime} \to s_{timeInterval}} $
    \item $\mathsf{instant : value \to s_{instantTime}}$
    \item $\mathsf{addEndCause : set[entity] \times s_{causality} \to s_{causality}}$

\end{itemize}
  $\varepsilon_\mathsf{C}, \varepsilon_\mathsf{S}, \varepsilon_\mathsf{P}, \varepsilon_\mathsf{A}$ are constants (0-ary functions) that represent the empty value for the sorts $s_\mathsf{causality}$,  $s_\mathsf{sequence}$, $\mathsf{s_{provenance}}$, $s_\mathsf{annotations}$  respectively.
 \end{example}
  These qualifier category terms can take different forms depending on the selected theories for the categories. For instance, the validity context is based here on temporal intervals specified by the \textsf{start time} and \textsf{end time} qualifiers only, but it could also be based on uncertain time intervals specified by some combination of \textsf{start time, end time, earliest start date, latest start date, earliest end date}, etc. This will determine the construction functions' arities for each sort. 
  

In Wikidata, the value $v$ of a statement may be specified as \emph{noValue}\footnote{https://www.mediawiki.org/wiki/Wikibase/DataModel}. It indicates that we know that the statement's subject is not connected to any value through the statement's predicate. In this case, the statement will be represented by an atom
$$ \mathsf{sno}(s, p, V, C, S, A, P) $$
of arity
$$\mathsf{entity} \times \mathsf{property} \times s_\mathsf{validity} \times s_\mathsf{causality} \times s_\mathsf{sequence} \times s_\mathsf{annotations} \times s_\mathsf{provenance}$$
A statement's value may also be specified as \emph{someValue} to indicate that there is some value for this predicate, although we don't know it. Such statements are represented by an atom
$$\mathsf{ssome}(s, p, V, C, S, A, P)$$
of arity
$$\mathsf{entity} \times \mathsf{property} \times s_\mathsf{validity} \times s_\mathsf{causality} \times s_\mathsf{sequence} \times s_\mathsf{annotations} \times s_\mathsf{provenance}$$
Qualifier values may also be 'noValue' or 'someValue'. This means that the $\mathsf{value}$ sort must have two additional constants: $\mathsf{noValue}$ and $\mathsf{someValue}$ and that the sort construction functions must take these values into account. 

\subsection{A basic algebraic specification for the Wikidata qualifier categories }

In this section, we present an algebraic specification for Wikidata. The specification contains six modules: \textsf{Validity, Causality, Sequence, Annotations, and Provenance}, one for each qualifier category shown in section 2. The goal is to show the methodology to couple the MSL with an algebraic specification for reasoning on qualifiers. Therefore the modules define only basic theories. Depending on the intended application domain, more sophisticated theories might be needed. For instance, a historical application would necessitate the introduction of uncertainty in the validity time theory (based on the qualifiers such as \textsf{earliest date}, \textsf{latest date}, \textsf{earliest start date}, etc.). Handling recurrent periods (summer, Monday, night, Christian Lent, etc.), indicated by qualifiers such as \textsf{valid in period}, would require yet another time theory.

The theories proposed in the basic specification have the following characteristics 

\begin{description}
     \item[Validity context] A validity context is a region with several dimensions in an abstract validity domain. In this specification, we work on validity time and validity space only. We develop a notion of certain time\footnote{The goal here is not to develop all the representation of time in Wikidata but to show the methodology to combine the many-sorted logic and the specification in the reasoning rules later.}. Moreover, we consider that an undefined validity means valid everywhere.   
    \item[Causality] The causality of a statement is a set of (beginning) causes and a set of end causes. Causes are propagated by inference: we consider that the causality of the inferred statement is the union of the causalities of the body statements.
    \item[Sequence] The sequence of a statement indicates the position of the subject in a sequence of items that have the same property and object.
    \item[Provenance] The provenance records the sources from which a statement was derived, the data acquisition techniques, etc. When two statements are used in an inference, their provenances are propagated by inference in the derived statement. They represent everything that was needed to derive the statement.
    \item[Annotations] An annotation is a set of attribute-value pairs. Most of them do not propagate through ontological inferences.
\end{description}

Table  \ref{tab:moduleop} shows the main operations defined in these modules. The full specification is in the appendix. 

\begin{table}[htbp!]
    \centering
\begin{tabular}{>{\sffamily}l>{\sffamily}l>{\sffamily}l>{\sffamily}l}

Module & Constructors & Accessors & Predicates \\
\hline \\
\textbf{Value} & value (IRI) & & equal \\
     & value (String) \\
     & ... \\
    & undefined \\
    
\hline \\
\textbf{ValidityContext} & $ \varepsilon_{V}$&        & testIntersectValidity \\
                & timeValidity(time)     & extractTime   & equal \\
                & spaceValidity(space)   & extractSpace        & includes \\
                & setTime            &  \\
                & setSpace \\
                & inter$_V$ \\
\hline \\
\textbf{Causality}  & $ \varepsilon_{C}$ &  getEndCause   & equal \\
                    & addHasCause       & getHasCause   & \\
                    & addEndCause       &   \\
                    & union$_C$         &  \\
                    & inverseCause      &  \\
\hline \\
\textbf{Sequence}    & seq(entity, entity)    & next         & hasNext \\
                     & seqWithNext(entity)   & previous     & hasPrevious \\
                     & seqWithPrev(entity) & \\
                     & $ \varepsilon_{S} $&  \\
\hline \\
\textbf{Annotations}     & $\varepsilon_{A}$   &              &  \\
(one add/get pair of     & add$A_1$            &  get$A_1$    &  \\
 operations for each     & ...                 &  ...         &  \\
 considered qualifier)   & add$A_n$            &  get$A_n$    &  \\
\hline \\
\textbf{Provenance} & $\varepsilon_{P}$   &  getSources       & equal \\
                    & addSources          &               & \\
                    &  union$_P$          &               &  \\
\hline \\
\end{tabular}
    \caption{Main operations of the sorts modules.}
    \label{tab:moduleop}
\end{table}
As an example, we present in more detail the causality module. 
The causality module is used to specify the behavior of the causality qualifiers in the reasoning. 
In the following, we present part of this module, notably the operation's signature. The complete module can be found in the specification. The syntax is CASL
\footnote{The operations \texttt{emptyValidity}, \texttt{emptyCause}, \texttt{emptySequence}, \texttt{emptyProvenance}, \texttt{emptyAnnnotations} are denoted as $\varepsilon_V$, $\varepsilon_C$, $\varepsilon_S$, $\varepsilon_P$, $\varepsilon_A$ respectively in the body of the article, for reasons of brevity. We also use subscripts for the function name, such as union$_C$, instead of \textsf{unionCause} because subscripts cannot be written in CASL. }  \cite{astesiano2002casl}. 
\begin{lstlisting}[
  mathescape,
  columns=fullflexible,
  basicstyle=\fontfamily{lmvtt}\selectfont,
  morekeywords={spec,then,sort,op,pred,forall},
  keywordstyle=\bfseries
]
spec Causality = Value then SET[sort entity] then

sort causality

%% generators
op emptyCause : causality
op addEndCause : set[entity] * causality -> causality 
op addHasCause : set[entity] * causality -> causality 

pred equal : causality * causality 
op getEndCause : causality -> set[entity]
op getHasCause : causality -> set[entity]
op unionCause : causality * causality -> causality

forall c1, c2: causality
. equal(c1, c1) <=> equal(getEndCause(c1), getEndCause(c2)) 
                    /\ equal(getHasCause(c1), getHasCause(c2))
. getEndCause(emptyCause) = emptyset
. getEndCause(addEndCause({e1}, c1)) = add(e1, getEndCause(c1))
. getEndCause(addHasCause({e1}, c1)) = getEndCause(c1)
. getEndCause(unionCause(c1, c2)) = getEndCause(c1) union getEndCause(c2)
. getEndCause(inverseCause(addEndCause({e1}, c1))) = add(inverseCause(e1), 
 getEndCause(inverseCause(c1)))
. getEndCause(inverseCause(addHasCause({e1}, c1))) = getEndCause(inverseCause(c1))

. . . 
\end{lstlisting}
Since the representation of a Wikidata statement includes an object for each type of sort (qualifier categories), the operation $\varepsilon_\mathsf{C}$ is in charge of generating an empty causality object. In addition, the value of the causality qualifiers can be multiple. To make it possible to add many causality resources, we use sets (e.g., set[resource]). Hence, the two functions \textsf{addEndCause} and \textsf{addHasCause} add a set of resources to the causality objects. Similarly to adding causality objects, we might want to get these objects, which explains the use of the functions \textsf{getEndCause} and \textsf{getHasCause}. Furthermore, the \textsf{equal} predicate is used to check if two statements have equal causalities. Finally, we introduced the  \textsf{unionCause} function to compound the causality of many statements.

\section{Handling qualifiers in reasoning }
This section shows how the MSL and the algebraic specification facilitate the reasoning on knowledge
graph qualifiers such as Wikidata. Instead of dealing with hundreds of qualifiers while reasoning, we work on a higher view of the qualifiers (i.e., the sorts) using the MSL, and we manipulate the lower level (i.e., qualifiers) using the algebraic specification. To exemplify the MSL +  Algebraic Specification methodology, we propose to study different cases: 
\begin{enumerate}
    \item  Inference rules for the "ontological" properties\footnote{Wikidata has no formal semantics and no entailment regime (i.e., there are no inference rules that we can use to show the utility of MSL).}.
    \item  Inference rules  based on the semantics of qualifiers (e.g., for sequence qualifiers). 
    \item Domain-specific inference rules.
\end{enumerate}
 
In each case, we consider MSL inference rules of the form $\varphi \longrightarrow  \psi $ where $\varphi $ is a conjunction of MSL atoms, and $\psi$ is an MSL atom. 


\begin{table}  [hbt]
\begin{tabular} { |p{4.2cm}| p{1.9 cm} |p{4.5cm}| }
 \hline
 \textbf{Wikidata ontological constructs} & \textbf{Prominence} &\textbf{RDFS/OWL property}  \\
 \hline
instance of (P31) & 102 229 515 & rdf:type     \\
\hline
 subclass of (P279) & 3 269 258 & rdfs:subClassOf   \\
\hline
 subproperty of (P1647) & 783 & rdfs:subPropertyOf   \\
\hline
\hline
 equivalent property (P1628) & 658 & owl:equivalentProperty     \\
\hline
 equivalent class (P1709)    & 1377 & owl:equivalentClass  \\
\hline
 inverse property (P1696)    & 171 & owl:inverseOf  \\
\hline
different from (P1889)    & 786 135 & owl:differentFrom  \\
\hline
\hline
value-type constraint (Q21510865) &1 025& rdfs:range \\
\hline
subject type constraint (Q21503250) & 5 973 & rdfs:domain \\
\hline
symmetric constraint (Q21510862) & 45 & owl:SymmetricProperty \\
\hline
inverse constraint (Q21510855) & 123 & owl:inverseOf \\
\hline
\end{tabular}
\caption{\label{tab:mapping  Wikidata OWL}Mapping of Wikidata ontological properties to RDFS/OWL properties.} 
\end{table}

\subsection{Ontological reasoning in Wikidata}
As mentioned previously, Wikidata contains some ontological properties\footnote{They are not axiomatized. They are like the other data in Wikidata.}. Table \ref{tab:mapping  Wikidata OWL} illustrates the prominence of the Wikidata ontological properties and the correspondence between them and RDFS/OWL properties. As we can see, some of them are declared as Wikidata properties and others as Wikidata constraints\footnote{\url{https://www.wikidata.org/wiki/Help:Property_constraints_portal}}. Like any other statement in Wikidata, statements containing ontological properties are also annotated with qualifiers. Different categories of qualifiers are used with different properties. For instance, in the Wikidata dump, we are working on, while the \textsf{subclass of} property can be annotated with all qualifier categories, the \textsf{subproperty} is annotated only with validity contexts, annotations and provenance qualifiers. Table \ref{tab:catePerQual} shows each ontological property with the different categories of qualifiers that can annotate it in Wikidata.
 We have studied the qualifier behavior for each ontological property/constraint to define a corresponding inference rule (when possible). Let us consider two of these rules (the others can be found in Appendix A).

\begin{table}[hbt]
\centering
 \begin{tabular}{| p{.40\textwidth} | p{.06\textwidth} |p{.06\textwidth}| p{.06\textwidth}| p{.06\textwidth}|p{.06\textwidth}|} 
\hline
   \textbf{Ontological construct} & \textbf{V} & \textbf{C} & \textbf{S}  & \textbf{A}  & \textbf{P} \\
 \hline
\textsf{instance of }(P31) & \Checkmark  & \Checkmark & \Checkmark  & \Checkmark & \Checkmark  \\
 \hline
\textsf{subclass of }(P279) & \Checkmark  & \Checkmark & \Checkmark  & \Checkmark & \Checkmark  \\
 \hline
 \textsf{subproperty of }(P1647) & \Checkmark  & &  & \Checkmark & \Checkmark  \\
 \hline
 \textsf{equivalent property  }(P1628) &  & &  &\Checkmark & \Checkmark  \\
 \hline
  \textsf{equivalent class  }(P1709) &  & &  & \Checkmark & \Checkmark  \\
 \hline

   \textsf{inverse of   }(P1696) &  & &  & \Checkmark &\Checkmark  \\
 \hline
    \textsf{different from  }(P1889) & \Checkmark  & \Checkmark&\Checkmark  & \Checkmark&\Checkmark  \\
 \hline

 \end{tabular}
\caption{\label{tab:catePerQual}The qualifier categories annotating each ontological property: V, C, S, A, P stand for validity contexts, causality, sequence, annotations and provenance, respectively.}
\end{table}
\paragraph{The \textit{instance of} rule.} Wikidata contains an  \textsf{instance of (P31)} property. It is similar to \textsf{rdf:type} which inspires the following inference rule:

$$\begin{array}{l}
\SP(x, \mathsf{instance\ of}, y, V_1, C_1, S_1, A_1, P_1) \\

\wedge \quad  \SP(y,\mathsf{subclass \ of} , z, V_2, C_2, S_2,A_2, P_2) \\
\wedge \quad \mathsf{testIntersectValidity}(V_1, V_2) \\
\longrightarrow \SP(x, \mathsf{instance\ of}, z, \mathsf{inter}_{V}(V_1, V_2), \mathsf{union}_{C}(C_1,C_2),
\varepsilon_{S} ,
\varepsilon_{A},
\mathsf{union}_{P}(P_1,P_2)
)
\end{array}$$
According to the \textsf{instance of} statistics\footnote{\url{http://ke.unige.ch/wikidata/Statistics/QualifiersByProperty/P31-qualifiers.csv}}, a statement containing \textsf{instance of}  is usually annotated with many  qualifiers belonging to different sorts, but mostly to the annotations sort. The MSL representation lets us represent the sorts instead of the qualifiers, making the representation compact. The body of this rule represents the fact that if $x$ is an instance of $y$ with some validity context $V_1$, causality $C_1$, sequence $S_1$, annotations $A_1$ and provenance $P_1$ and $y$ is a subclass of $z$  with some validity context $V_2$, causality $C_2$, sequence $S_2$ and provenance $ P_2$, and there is an intersection between $V_1$ and $V_2$;  then $x$ is an instance of $z$  with:
\begin{itemize}
    \item A validity context that is the intersection of the two validity context statements (i.e.,  $V_1$ and $V_2$). We use the $\mathsf{inter}_{V}$ from the module validity context of the specification\footnote{If V2 does not provide temporal qualifiers, the \textsf{testIntersectValidity} function will return true because it is considered that V2 is not temporally constrained (i.e. covering all times).}.

    \item A causality that is the union of the two causalities (i.e. $C_1$ and $C_2$). We use the $\mathsf{union}_{C}$ function of the causality module.
    \item An empty sequence  $\varepsilon_{S}$ because the ordering of $x$ among the instances of $y$, if it exists, cannot serve to infer the ordering of $x$ among the instances of the superclass $z$. There may not even be an ordering in $z$. For instance, New Mexico has \textsf{ordinal number} 47 as an instance of \textsf{U.S. State} but does not have any ordinal number in the superclass \textsf{state}.
    \item An empty annotation $\varepsilon_{A}$ because a close examination of the annotations qualifiers used with \textsf{instance of} and \textsf{subclass of} in wikidata shows that they cannot be copied.
    \item A provenance that is the union of the two provenances (i.e., $P_1$ and $P_2$). We use the $\mathsf{union}_{P}$ function of the provenance module.
\end{itemize}

\paragraph{The symmetry rule}
Inspired from the \emph{Symmetric property} constraint in Wikidata  (\textsf{Q18647518}). Constraints in Wikidata express regularities (patterns of data) that should hold in general\footnote{In practice, they identify constraint violations to contributors who can either fix the problem or determine that the particular anomaly is acceptable.}. The declaration that a property $p$ has a Wikidata constraint $c$ is done with a statement \textsf{st($p$, property constraint,  $c$, $\varepsilon_{V}$,  $\varepsilon_{C} $,  $\varepsilon_{S} $,  $A_0$, $P_0$)}. The sorts in play are only annotations and provenance. Most of the annotations qualifiers describe the constraint.

The symmetric constraint\footnote{\url{https://www.wikidata.org/wiki/Help:Property_constraints_portal/Symmetric}} specifies that a property is symmetric, and values for that property should have a statement with the same property pointing back to the original item. It gives rise to an inference rule.

$$\begin{array}{l}
\SP(p, \textsf{property constraint}, \textsf{symmetric property}, \varepsilon_{V}, \varepsilon_{C}, \varepsilon_{S}, A_0, \varepsilon_{P}  ) \\
\wedge \quad  \SP(x, p, y, V_1, C_1, S_1, A_1, P_1) \\
\longrightarrow \SP(y, p, x, V_1, \mathsf{inverseCause}(C_1), \varepsilon_{S}, \varepsilon_{A}, P_1)
\end{array}$$
The body of this rule represents a statement with a property $p$ that is symmetric, a validity context $V_1$, causality $C_1$, the sequence $S_1$, annotations $A_1$ and provenance $P_1$. The inferred statement contains:
\begin{itemize}
    \item A validity context that is the validity context of the original statement because the validity time of the statement will remain the same in both directions of the property (i.e., $\SP(x, p, y, ...)$ or $\SP(y, p, x, ...)$).
    \item For the causality, the situation is slightly different: most of the causality values are preserved (e.g., $\mathsf{divorce}$ or $\mathsf{death}$). However, others must be replaced by their ``inverse'' (e.g. $\mathsf{death\ of\ subject}$ $\to$ $\mathsf{death\ of\ object}$) or dropped. This is accomplished by the $\mathsf{inverseCause}$ function.
    \item An empty sequence  $\varepsilon_{S}$. The following Wikidata statement shows why:
\begin{equation}
\begin{array}{l}
 (\textsf{Hino Yasuko, spouse, Ashikaga Yoshimitsu})[ \\
\quad\textsf{replaces}: \textit{Hino Nariko}, \\ 
] \\ 
\end{array}
\end{equation}
As we can see, the sequence qualifier \textsf{replaces} cannot be preserved for the inferred statement because it relates to the subject  (i.e., Hino Yasuko replaces Hino Nariko).
\item  An empty annotation $\varepsilon_{A}$.
\item  A provenance $P_1$ that is the provenance of the original statement.
\end{itemize}

\subsection{Inference rules induced by qualifier semantics}
In addition to the above-mentioned ontological rules, the Wikidata model has specific features that can give rise to additional rules.  Some of  these inference rules come from the  semantics of qualifiers,  such as the sequence qualifiers. 
\paragraph{The \textit{sequence previous} rule} Inspired from the \textit{replaces(P1365)}/ \textit{follows(P155)} qualifiers in Wikidata. 
$$\begin{array}{l}
\SP(x, p, y,V_1, C_1, S_1, A_1, P_1) \land  \mathsf{hasPrevious}(S_1) \\
\longrightarrow \\
\SP(\mathsf{previous}(S_1), p, y,\\
    \qquad\mathsf{setTime}(V_1, \mathsf{interval}(\mathsf{undefined}, \mathsf{startTime}(\mathsf{extractTime}(V_1))),\\
    \qquad\varepsilon_{C}, \mathsf{seqWithNext}(x), \varepsilon_{A}, P_1) \\
\end{array}$$
This rule checks if there is a sequence qualifier of type "previous" in a statement (i.e., \textsf{replaces}). If the statement has a sequence qualifier of type previous, which will be detected using the $\mathsf{hasPrevious}(S_1)$  then in the conclusion of the rule, we generate a new statement where:
\begin{itemize}
    \item The validity context of this statement is constructed using the $\mathsf{setTime}$ function of the validity context module. In this case, this function takes as input: $V_1$ for the first input and the $\mathsf{interval}$ function from the validity time module for the second input. The $\mathsf{interval}$  function has \textsf{undefined}\footnote{undefined means unconstrained} as the start time because we cannot infer anything for the start time but the end time is equal to the validity context's start time of the original statement $V_1$; this is achieved by using the $\mathsf{startTime}(\mathsf{extractTime}(V_1))$ function.
    \item We cannot infer a causality in this case: we cannot copy the original statement causality, nor infer a new one.
    \item The sequence qualifier of the inference is the subject of the original statement $x$. Since the qualifier that generated this inference is of type previous, then the sequence qualifier of the generated inference will be of type next and this is illustrated using the function $\mathsf{}seqWithNext(x)$ of the sequence module.
    \item We keep the provenance of the original statement $P_1$.
\end{itemize}

\subsection{Domain specific rules}

Wikidata covers a lot of knowledge domains which gives rise to domain rules. For example, we find the \textsf{spouse} property used to say that the subject has the object as their spouse (e.g., husband, wife, partner, etc.).

\begin{example}\label{ex:marriage and death}

If the end of the validity of a marriage coincides with the death date of the subject spouse, one can infer that the death caused the end of the marriage.

$$\begin{array}{l}
\SP(x_1, \mathsf{spouse}, y_1,  V_1, C_1, S_1, A_1, P_1) \\
\land \quad  \SP (x_1, \mathsf{date \ of \ death},d, V_2, C_2, S_2, A_2, P_2) \\ 
\land \quad \mathsf{equal}(\mathsf{instant}(d), \mathsf{endTime}(\mathsf{extractTime}(V_1)))\\
\longrightarrow 
\SP (x_1,  \mathsf{spouse},
y_1,  V_1, \mathsf{addEndCause}(\mathsf{death\ of\ subject}, C_1), S_1, A_1, \mathsf{union}(P_1, P_2))
)
\end{array}$$
\end{example}
The body of the rule contains the statement $\SP(x_1, p, y_1,  V_1, C_1, S_1, A_1, P_1)$, the death statement $\SP(x_1, \mathsf{date of death},$ $date, V_2, C_2, S_2, A_2, P_2)$ and the condition on the death date. The inferred statement contains the same subject, predicate and value, the same validity, sequence, annotation, and provenance, but :
\begin{itemize}
    \item A new causality where \textsf{death of subject} has been added.
    \item A provenance that is the union of $P_1$ and $P_2$
\end{itemize}
 
\section{Implementation}
This section shows the methodology and tools we developed to implement MSL reasoning on (a subgraph of) an RDF dump of Wikidata. It consists of the following steps:

\subsection*{Sort operation specification}
This first step must produce an MSL vocabulary and an algebraic specification to determine the sort operation semantics. This can be done by re-using or extending the specification in Section 4 and Appendix B, depending on the application domain. Currently, the specification must be written in CASL and at least parsed and statically analyzed with the HETS toolset\footnote{\url{http://hets.eu/}}. Extending the specification is further described in Section \ref{sec:extens}.

\subsection*{Sort value representation}
This step aims to define a representation of the sort values as RDF literals. Such a representation simplifies the handling and processing of the sort values in the RDF graph, SPARQL queries, and Javascript functions. In particular, new values can be generated without creating new IRIs or blank nodes in the graph. The values can be passed to and returned by external Javascript functions as one object. Using JSON strings eases the encoding/decoding and manipulation of values. For example, a causality value can be represented as
\begin{lstlisting}[
  backgroundcolor = \color{lightgray},
  mathescape,
  columns=fullflexible,
  basicstyle=\fontfamily{lmvtt}\selectfont,
  morekeywords={spec,then,sort,op,pred,forall,end},
  keywordstyle=\bfseries
]
    {"hascause": [ ], "endcause": [wd:Q796919]}
\end{lstlisting}
where \texttt{wd:Q796919} is the IRI representing the Wikidata entity \textsf{resignation} in a RDF dump.

\subsection*{Implementation of the sort operations}
The sort operations defined in the specification must be implemented as JavaScript functions. This choice is motivated by the fact that many triple stores allow users to define JavaScript functions that can be invoked in SPARQL queries. Since the sort values are JSON strings, the functions typically follow the pattern:
    \begin{lstlisting}[
     backgroundcolor = \color{lightgray},
     mathescape,
  columns=fullflexible,
  basicstyle=\fontfamily{lmvtt}\selectfont,
  morekeywords={function, var},
  keywordstyle=\bfseries
]
    function op(string_p1, string_p2, ...){
      var p1 = JSON.parse(string_p1)
      var p2 = JSON.parse(string_p2)
      ...
      ... operation computation
      ...
      result = ...
      string_result = JSON.stringify(result)
      return string_result
\end{lstlisting}

The algebraic specification provides proof obligations for each functions as well as skeletons for generating unit tests.

\subsection*{Generating the sort values}

 In an RDF dump of a Wikidata subgraph, each statement is represented by a node that is connected through triples to the statement subject, value, and qualifier values. The predicate of these triples are the property and qualifier names of the statement. For example, the statement about the marriage (\textsf{P26)}) of George C. Scott (\textsf{Q182450}) and Colleen Dewhurst (\textsf{Q253916}) (Section 1, statement (1)) is represented by a statement IRI (\textsf{wds:Q182450\--3A25317F\--3088\--4113-\-8D5A-52\-375AB21FAE}) that participates in the triples.

 \begin{lstlisting}[
 backgroundcolor = \color{lightgray},
  mathescape,
  columns=fullflexible,
  basicstyle=\fontfamily{lmtt}\selectfont,
  morekeywords={function, var},
  keywordstyle=\bfseries
]
wd:Q182450 p:P26 wds:Q182450-3A25317F-3088-4113-8D5A-52375AB21FAE .
wds:Q182450-3A25317F-3088-4113-8D5A-52375AB21FAE ps:P26 wd:Q253916 ;
    pq:P580 "1960-01-01T00:00:00Z"^^xsd:dateTime ;                     $\mathit{\ (start\ time)}$
    pq:P582 "1965-01-01T00:00:00Z"^^xsd:dateTime ;                     $\mathit{\ (end\ time)}$
    pq:P1534 wd:Q93190.                                                $\mathit{\ (cause\ \ divorce)}$
\end{lstlisting}
 
 The role of the sort generation step is to compute the sort values that represent the statement's qualifiers and to connect them to the statement through triples of the form. 
 
 $$\mathit{statement\ IRI}\ \ \ \mathsf{pq{:}}\mathit{sort\ name} \ \ \ \verb+"+\mathit{sort\ value\ in\ JSON}\verb+"+$$
The sort values are obtained by applying the sort constructor functions to the qualifier values.  For example, the sorts of the statement shown in Example \ref{ex:scott} would be represented as

\begin{lstlisting}[
 backgroundcolor = \color{lightgray},
  mathescape,
  columns=fullflexible,
  basicstyle=\fontfamily{lmtt}\selectfont,
  morekeywords={function, var},
  keywordstyle=\bfseries
]
wds:Q182450-3A25317F-3088-4113-8D5A-52375AB21FAE
  pq:validityJ
    '{"time": {"start":"1960-01-01T00:00:00Z",
               "end":"1965-01-01T00:00:00Z"},
      "space":{}}' ;
  pq:causalityJ '{"hasCause":[],"endCause":["wds:Q93190"]}' ;   $\mathit{\ \ (divorce)}$
  pq:sequenceJ : '{}' ;
  pq:annotationsJ : '{}'
  pq:provenanceJ : '{}'
\end{lstlisting}

The sort value generation is carried out by generation tool with user-definable generation functions for each sort.

\subsection*{Define and execute the inference rules}

We defined a concrete syntax for the MSL rules. For example, the rule of Example \ref{ex:marriage and death} is written as

\begin{lstlisting}[
 backgroundcolor = \color{lightgray},
  mathescape,
  columns=fullflexible, keepspaces=true, 
  basicstyle=\fontfamily{lmtt}\selectfont,
  morekeywords={function, var},
  keywordstyle=\bfseries
]
st(X1, :P26, Y1, V1, C1, S1, A1, P1)     %% X1 spouse Y1
st(X1, :P570, D, V2, C2, S2, A2, P2)     %% X1 date of death D
equal(D, endTime(extractTime(V1)))
->
st(X1, :P26, Y1, V1, addEndCause(:Q99521170, C1),  %% death of subject
   S1, A1, P1)
\end{lstlisting}

A rule compiler translates these rules into SPARQL CONSTRUCT queries. The compiled rules are then executed by a naive rule engine that repeatedly runs all the rules and adds the generated triples to an inferred graph until nothing new can be generated.


The code of the prototype implementation can be found at \url{http://ke.unige.ch/wikidata/}.



\section{Extensibility}
\label{sec:extens}
This work is a basis for many possible extensions concerning the axiomatization of the qualifier categories and the corresponding reasoning. We choose to use algebraic specification because the possible extension of the existing specification is part of the general theoretical framework of this logical model. 
However, due to the universal scope of axioms in logic, constraining algebraic specification to respect already-defined properties requires careful examination of the impact of extensions. 
Flattening the specification (which amounts to putting the whole specification in a single module) is generally not a good approach as the modular structure of the specification is essential, and already existing reasoning is not supposed to be modified on the ground terms. In an algebraic specification framework, it is possible to generate proof obligations for existing parts that are extended, allowing to (automatically) check the quality of the extensions in terms of consistency and completeness and absence of perturbation on the base specifications. 


\begin{example}

Suppose we add 'noValue' as a particular causality situation not considered in the original specification. In that case, we must carefully extend all the axioms already defined in the \textsf{Causality} module by this value. Sometimes, it is necessary to differentiate new behaviors through specific axioms. These axioms must deal with the new behaviors induced by new extension values for consistency and completeness reasons. 

\begin{lstlisting}[
  mathescape,
  columns=fullflexible,
  basicstyle=\fontfamily{lmvtt}\selectfont,
  morekeywords={spec,then,sort,op,pred,forall},
  keywordstyle=\bfseries
]
spec CausalitywithnoValue = Causality then

%% generators
op noValue : causality

forall e1 : entity, c1, c2, c3 : causality,se1: set[entity]

. unionCause(noValue, c1) = c1
. unionCause(c1, noValue) = c1
. addEndCause(se1, noValue) = ...
. addHasCause(se1, noValue) = ...



\end{lstlisting}

\end{example}

\begin{example}

The current algebraic specification handles only two dimensions in a validity context: validity time and validity space. However, Wikidata covers different domains, which may have their specific validity contexts. For instance, Beghaeiraveri et al.  \cite{beghaeiraveri2021experiences} extract six Wikidata subsets corresponding to six Wikidata Wikiprojects: Gene Wiki, Taxonomy, Astronomy, Music, Law, and Ships. 
\begin{itemize}
    \item In biology, a taxon (group of organisms) can be considered as a validity context: The \textsf{median lethal dose (LD50)} of a chemical compound is valid only for a particular taxon. 
    \item In the arts, a work can be a validity context: The statement (\textsf{USS Enterprise}, \textsf{crew members(s)}, \textsf{Arex Na Eth}) is valid only in work \textsf{Star Trek: The Animated Series}.
\end{itemize}

These dimensions can be taken into account by extending the \texttt{Validity} specification. Some validity dimensions are orthogonal to space and time (e.g., the taxon dimension and the work dimension are independent of time and space). In this case, the extension is straightforward as it does not perturb the existing specification.

\end{example}
\section{Related works}
Reasoning with Wikidata qualifiers was discussed in a paper entitled \emph{Logic on MARS} \cite{marx2017logic}. The authors give a formalization of a generalized notion of property graphs, called multi-attributed relational structures (MARS), and introduce a matching knowledge representation formalism, multi-attributed predicate logic (MAPL). The formalism could be applied to multi-attributed knowledge graphs like Wikidata. A decidable fragment of MAPL was then proposed. Several useful expressive features characterize the fragment, among them the class of formulae used to express conditions on attribute-value sets. The fragment also includes functions. In addition, they introduce a rule-based fragment of MARPL, where rules are allowed to contain arbitrary specifiers. A couple of years later, MARS was extended with Wikidata data types \cite{patel2020wikidata}. Their work supports the explicit expression of Wikidata's ontological axioms and provides a means for handling them. In an extended technical report \cite{martin2020logical}, the authors also explain how they support the expression of nearly all current Wikidata property constraints, plus various other constraints.  Our work differs from MARS and its extension because in MARS: 1) the authors treat the value of qualifiers as opaque, i.e., they do not provide special operations or reasoning services for qualifiers types, such as time, causality, provenance etc., 2) they handle validity and additive qualifiers in a uniformed manner 3) they work directly on the qualifiers which is cumbersome if a statement is annotated with a lot of them. In our work, we explore another line of research where: 1) we handle different categories of qualifiers (i.e., the nature of qualifiers is not opaque), 2) we work on the sorts values instead of the qualifiers.

If we discuss the topic from a broader view, the state of the art of contextual knowledge representation and reasoning is large and dates from the late 1980s when John McCarthy proposed to formalize contexts as a crucial step toward solving the problem of generality. \cite{mccarthy1987generality}. Some theoretical works have been proposed, such as the two-dimensional description logics \cite{klarman2011two} with a core dimension to describe object knowledge and a context dimension to describe object knowledge. The work was later adopted to build OWL$^{C}$ \cite{aljalbout2019owl}, which extends the OWL2 RL profile and adds validity contexts to all inference rules. These works propose solutions to deal only with validity context.  Other works \cite{zimmermann2012general} deal with a  finite number of contexts/annotations such as time, provenance, and trust. They use lattice structures to express complex computations on contexts.  
In fact, the presence of intersection operations in our algebraic specification also shows that some qualifier categories (sorts) form semi-lattices (equipped with additional operations). However, this is not the case for every sort. Moreover, the lattice meet and join operations are not always sufficient to express the contexts in the inferred statements.  


\section{Conclusion and Future Works}
This paper tackled the challenging topic of finding a model for qualified knowledge graphs in general, and  Wikidata in particular, that proposes a unified/structured way to represent the variety of statement qualifiers and to incorporate them into inference rules. First, we proposed a categorization of the massive number of qualifiers. The most prominent qualifiers fall into the following categories: validity context, causality, sequence, annotations, and provenance. Using this categorization, we formalized the representation of Wikidata statements within a many-sorted logic coupled with an algebraic specification. The sorts of the logic represent the different categories of qualifiers. The algebraic specification contains the functions that are applied to qualifiers while reasoning. Using this logic and specification, we demonstrated that most of the Wikidata qualifiers can be formalized in reasoning. In the last part of the paper, we explained the methodology for implementing the MSL approach. Finally, we showed how the approach can be extended to cover other categories of qualifiers and different theories of qualifier categories. In future works, we plan to show more of the extensibility of the approach by implementing various use cases related to the variety of topics Wikidata covers. We plan to conduct practical experiments on Wikidata subsets that correspond to well-defined domains, i.e, to develop comprehensive sets of rules for these domains, execute them and evaluate the quality and completeness of the inferred statements as well as the efficiency of the process. Furthermore, we plan to show reasoning on Wikidata constraints using the constraint subcategory of the annotations. Since the annotation category contains a great diversity of qualifiers, we also plan to study possible sub-classifications of these qualifiers further and specify the corresponding theories.







\bibliography{bibliography}        

%

\begin{appendices}
\section{Ontological rules}
This appendix contains the rest of the ontological rules.
\paragraph{The \textit{subclass of} rule} Inspired from the \textit{subclass of (P279)} property in Wikidata.\newline\newline
$\begin{array}{l}
\SP(x, \mathsf{subclass \ of}, y, V_1, C_1, S_1, A_1, P_1) \\
\wedge \quad  \SP(y, \mathsf{subclass \ of}, z, V_2, C_2, S_2, A_2, P_2) \\
\wedge \quad (\mathsf{testIntersectValidity}(V_1, V_2)) \\
\longrightarrow \SP(x, \mathsf{subclass \ of}, z, \mathsf{inter}_{V}(V_1, V_2), \mathsf{union}_{C}(C_1,C_2),
\varepsilon_{S},
\varepsilon_{A},
\mathsf{union}_{P}(P_1,P_2)
)
\end{array}$
\paragraph{The \textit{subproperty of} rule} Inspired from the \textit{subproperty of (P1647)} in Wikidata.\newline \newline
$\begin{array}{l}
\SP(p, \mathsf{subproperty \ of}, q, V_1, \varepsilon_{C}, \varepsilon_{S}, A_1, P_1) \\
\wedge \quad  \SP(q, \mathsf{subproperty \ of}, r, V_2, \varepsilon_{C}, \varepsilon_{S}, A_2, P_2) \\
\wedge \quad (\mathsf{testIntersectValidity}(V_1, V_2)) \\
\longrightarrow \SP(p, \mathsf{subproperty \ of}, r, \mathsf{inter}_{V}(V_1, V_2), \varepsilon_{C},
\varepsilon_{S},
\varepsilon_{A},
\mathsf{union}_{P}(P_1,P_2)
)
\end{array}$ \newline \newline \newline
$\begin{array}{l}
\SP(p, \mathsf{subproperty \ of}, q, V_1, \varepsilon_{C}, \varepsilon_{S}, A_1, P_1) \\
\wedge \quad  \SP(x, p, y, V_2, C_2, S_2, A_2, P_2) \\
\wedge \quad (\mathsf{testIntersectValidity}(V_1, V_2)) \\
\longrightarrow \SP(x, q, y, \mathsf{inter}_{V}(V_1, V_2), C_2,
S_2,
A_2,
\mathsf{union}_{P}(P_1,P_2)
)
\end{array}$

\paragraph{The \textit{different from } rule} Inspired from the \textit{different from (P1889)} property in Wikidata.  This property says that an item is  different from another item, with which it may be confused. 
\newline\newline
$\begin{array}{l}
\SP(p, \mathsf{different \ from}, q, V_1, C_1, S_1, A_1, P_1) \\
\longrightarrow \SP(q, \mathsf{different \ from}, p, V_1, C_1, S_1, A_1,
\mathsf{P_1}
)
\end{array}$


\paragraph{The \textit{Inverse property} rule:}  Inspired from the \textit{inverse property (P1696)} predicate in Wikidata . It means that the property has an inverse property, and values for the property should have a statement with the inverse property pointing back to the original item \footnote{Wikidata contains also an inverse constraint (Q21510855)} \newline\newline
$\begin{array}{l}
\SP(p, \textsf{inverse property }, q, \varepsilon_{V}, \varepsilon_{C}, \varepsilon_{S}, A_0, \varepsilon_{P}  ) \\
\wedge \quad  \SP(x, p, y, V_1, C_1, S_1, A_1, P_1) \\
\longrightarrow \SP(y, p, x, V_1, C_1, \varepsilon_{S}, \varepsilon_{A}, P_1)
\end{array}$ \newline\newline
\paragraph{Subject type constraint (Q21503250)} \textit{ This constraint  \footnote{\url{https://www.wikidata.org/wiki/Help:Property_constraints_portal/Subject_class}} is used to specify that items with the specified property must have one among a given set of types}

This constraint is similar to the  \textsf{rdfs:domain} property in RDFS. The qualifiers that can qualify this constraint belong to the annotations and provenance sorts. The qualifiers \textsf{class (P2308)} and \textsf{relation (P2309)} play an important role in the formalization of this constraint. They indicate that \textsf{relation} must hold between the subject of a statement with this property and one of the given classes. The \textsf{relation (P2309)} is either \textsf{instance of}, \textsf{subclass of} or \textsf{instance or sub-class of}. Note that the identifier used for \textsf{instance of} is wd:Q21503252, which is surprisingly different from (P31), the identifier of the conceptual \textsf{instance of} property. Similarly, for \textsf{subclass of}, the identifier used is "wd:Q21514624" and it is completely different from the identifier of the conceptual Wikidata \textsf{subclass of} (P279). In the following, we formalize this Wikidata constraint only when the \textsf{relation (P2309)} is equal to \textsf{instance of} (wd:Q21503252) or \textsf{subclass of}(wd:Q2151462). We do not formalize the \textsf{instance or sub-class of} because it is unclear what to do with it.

Since this constraint is disjunctive (the item must belong to at least one of the given classes) it cannot lead to a simple Horn rule. It must be represented as an existential rule (with the rare exception of a constraint whose class set is a singleton). The same is true for the value type constraints. \newline\newline

$\begin{array}{l}
\forall x, p, y, V_1, C_1, S_1, A_0, A_1, P_0,P_1, \exists T \\
 \textsf{\SP($p$, property constraint, type constraint, $\varepsilon_{V}$, $\varepsilon_{C}$, $\varepsilon_{S}$, $A_0$, $P_0$}) \\
\wedge \quad  \SP (x,p,y, V_1, C_1, S_1, A_1, P_1) \\
\wedge \quad \mathsf{contains}(\mathsf{getRelation}(A_0), \mathsf{" instance \ of ")}  \\
\longrightarrow
\SP(x, \textsf{instance of}, T, \varepsilon_{V}, \varepsilon_{C}, \varepsilon_{S}, \varepsilon_{A}, \mathsf{union}(P_0, P_1)) \wedge \mathsf{contains}(\mathsf{getClass}(A_0), T) 
\end{array}$ \newline \newline
Another form of the MSL formalization is possible when the \textsf{getRelation} predicate is \textsf{subclass of }.  \newline \newline
$\begin{array}{l}
\forall x, p, y, V_1, C_1, S_1, A_0, A_1, P_0,P_1, \exists T \\
 \textsf{
    \SP($p$, property constraint, type constraint, $\varepsilon_{V}$, $\varepsilon_{C}$, $\varepsilon_{S}$, $A_0$, $P_0$)
  } \\
\wedge \quad  \SP (x,p,y, V_1, C_1, S_1, A_1, P_1) \\
\wedge \quad \mathsf{contains}(\mathsf{getRelation}(A_0), \mathsf{" subclass \ of ")} \\
\longrightarrow
\SP(x, \textsf{subclass of}, T, \varepsilon_{V}, \varepsilon_{C}, \varepsilon_{S}, \varepsilon_{A}, \mathsf{union}(P_0, P_1)) \wedge \mathsf{contains}(\mathsf{getClass}(A_0), T)
\end{array}$\newline

\paragraph{Value type constraint (Q21510865)} \textit{The referenced item should be a subclass or instance of one among a given set of types}

This property is similar to the  \textsf{rdfs:range} property in RDFS.
The qualifiers that can qualify this constraint are 10:\textsf{ \{exception to constraint (P2303), constraint clarification (P6607), group by (P2304), class (P2308), relation (P2309), constraint status (P2316),  wikibase:rank, wikibase:hasViolationForConstraint, prov:wasDerivedFrom, rdf:type)\}}. They belong to the annotations and provenance sorts. The MSL formalization of this constraints takes two forms: \newline \newline

$\begin{array}{l}
\forall x, p, y, V_1, C_1, S_1, A_0, A_1, P_0,P_1, \exists T \\
 \textsf{\SP($p$, property constraint, value type constraint, $\varepsilon_{V}$, $\varepsilon_{C}$, $\varepsilon_{S}$, $A_0$, $P_0$}) \\
\wedge \quad  \SP (x,p,y, V_1, C_1, S_1, A_1, P_1) \\
\wedge \quad \mathsf{contains(getRelation(A_0), " instance \ of ") } \\
\longrightarrow
\SP(y, \textsf{instance of}, T, \varepsilon_{V}, \varepsilon_{C}, \varepsilon_{S}, \varepsilon_{A}, \mathsf{union}(P_0, P_1))\wedge \mathsf{contains(getClass(A_0), T)}
\end{array}$ \newline\newline

$
\begin{array}{l}
\forall x, p, y, V_1, C_1, S_1, A_0, A_1, P_0,P_1, \exists T \\
 \textsf{\SP($p$, property constraint, value type constraint, $\varepsilon_{V}$, $\varepsilon_{C}$, $\varepsilon_{S}$, $A_0$, $P_0$}) \\
\wedge \quad  \SP (x,p,y, V_1, C_1, S_1, A_1, P_1) \\
\wedge \quad \mathsf{contains(getRelation(A_0), " subclass \ of ")} \\
\longrightarrow
\SP(y, \textsf{subclass of}, T, \varepsilon_{V}, \varepsilon_{C}, \varepsilon_{S}, \varepsilon_{A}, \mathsf{union}(P_0, P_1))\wedge \mathsf{contains(getClass(A_0), T)}
\end{array}
$

\paragraph{Sequence Rules}
 \paragraph{The \textit{sequence next} rule} Inspired by the \textit{replaced by (P1366)}/ \textit{followed by(P156)} predicate in Wikidata.  the treatment of the qualifiers is analogous to the sequence previous rule. \newline \newline

$\begin{array}{l}
\SP(x, p, y, V_1, C_1, S_1, A_1, P_1) \wedge \quad  \mathsf{hasNext}(S_1) \\
\longrightarrow \\
\SP(\mathsf{next}(S_1), p, y,\\
    \qquad\mathsf{setTime}(V_1, \mathsf{interval}(\mathsf{endTime}(\mathsf{extractTime}(V_1)), \mathsf{undefined} ),\\
    \qquad\varepsilon_{C}, \mathsf{seqWithPrevious}(x), \varepsilon_{A}, ,P_1)
\end{array}$
 \section{Algebraic specification in CASL}
 \begin{lstlisting}[
  mathescape,
  columns=fullflexible,
  basicstyle=\fontfamily{lmvtt}\selectfont,
  morekeywords={spec,then,sort,op,pred,forall,end},
  keywordstyle=\bfseries
]
 spec NAT =

 sort nat

 op __+__ : nat * nat -> nat
 op __-__ : nat * nat -> nat
 pred __<__ : nat * nat

end

spec Value =

sort value
sort entity < value
sort datavalue < value
sort property < entity
sort IRI
sort string

%% generators
op value : IRI -> value
op value : string -> value
op entity : IRI -> entity
op property : IRI -> property
op datavalue : string -> datavalue   %% datavalue from a literal string


pred equal: value * value 

op undefined : value

forall v1, v2, v3: value, i:IRI
. equal(v1, v1) 
. equal(v1, v2) /\ equal(v2, v3) => equal(v1, v3)
. equal(v1, v2) => equal(v2, v1)

. not equal(value(i), undefined) 
. not equal(undefined , value(i))

op min : datavalue * datavalue ->? datavalue
op max : datavalue * datavalue ->? datavalue
op + : datavalue * datavalue ->? datavalue
op - : datavalue * datavalue ->? datavalue

end

spec TimeParam =

sort time
op undefined : time
op union : time * time -> time
op inter : time * time -> time

pred testIntersect : time * time 
pred incl : time * time 
pred equal : time * time

forall t1, t2, t3 : time
. equal(t1, t1) 
. equal(t1, t2) /\ equal(t2, t3) => equal(t1, t3)
. equal(t1 , t2) => equal(t2, t1) 

end


spec SpaceParam =

sort space

pred equal : space * space
pred incl : space * space
op union : space * space -> space
op inter : space * space -> space


forall s1, s2, s3 : space
. equal(s1, s1) 
. equal(s1, s2) /\ equal(s2, s3) => equal(s1, s3) 
. equal(s1, s2) => equal(s2, s1)
end


spec ValidityContext[TimeParam][SpaceParam] = 


sort validityContext


%% Generators 
op emptyValidity : validityContext %% empty means valid in every context 
op timeValidity : time -> validityContext
op spaceValidity : space -> validityContext
op timespaceValidity : time * space -> validityContext

pred testIntersectValidity : validityContext * validityContext 
pred incl : validityContext * validityContext 
pred equal : validityContext * validityContext 

op union : validityContext * validityContext -> validityContext 
op interValidity : validityContext * validityContext -> validityContext
op extractTime : validityContext -> time
op extractSpace : validityContext -> space
op setTime : validityContext * time -> validityContext
op setSpace : validityContext * space -> validityContext


forall c, c1, c2, c3 : validityContext, t, t1 , t2 : time , s, s1, s2 : space 
. setTime(emptyValidity, t) = timeValidity(t)
. setTime(timeValidity(t1), t2) = timeValidity(t2)
. setTime(spaceValidity(s), t) = timespaceValidity(t, s)
. setTime(timespaceValidity(t, s), t2) = timespaceValidity(t2, s)

. setSpace(emptyValidity, s) = spaceValidity(s)
. setSpace(timeValidity(t), s) = timespaceValidity(t, s)
. setSpace(spaceValidity(s), s2) = spaceValidity(s2)
. setSpace(timespaceValidity(t, s), s2)= timespaceValidity(t, s2)

. extractTime(timeValidity(t1)) = t1
. extractTime(timespaceValidity(t, s))= t
. extractTime(spaceValidity(s)) = undefined 

. extractSpace(spaceValidity(s)) = s
. extractSpace(timespaceValidity(t, s)) = s

. union(emptyValidity, c) = emptyValidity
. union(c, emptyValidity) = emptyValidity

. union(timeValidity(t1), spaceValidity(s2)) = emptyValidity 
. union(timeValidity(t1), timeValidity(t2)) = timeValidity(union(t1, t2))
. union(timeValidity(t1), timespaceValidity(t2, s2)) = timespaceValidity(union(t1, t2), s2)

. union(spaceValidity(s1), spaceValidity(s2)) = spaceValidity(union(s1, s2))
. union(spaceValidity(s1), timeValidity(t2)) = emptyValidity
. union(spaceValidity(s1), timespaceValidity(t2, s2)) = timespaceValidity(t2 , union(s1, s2))

. union(timespaceValidity(t1, s1), timespaceValidity(t2, s2)) = 
 timespaceValidity(union(t1, t2) , union(s1, s2))
. union(timespaceValidity(t1, s1), spaceValidity(s2)) = timespaceValidity(t1 , union(s1, s2))
. union(timespaceValidity(t1, s1), timeValidity(t2)) = timespaceValidity(union(t1, t2), s1) 


. interValidity(emptyValidity, c) = c
. interValidity(c, emptyValidity) = c

. interValidity(timeValidity(t1), spaceValidity(s2)) = timespaceValidity(t1, s2)
. interValidity(timeValidity(t1), timeValidity(t2)) = timeValidity(inter(t1, t2)) 
. interValidity(timeValidity(t1), timespaceValidity(t2, s2)) = timespaceValidity(inter(t1, t2), s1)
. interValidity(spaceValidity(s1), spaceValidity(s2)) = spaceValidity(inter(s1, s2)) 
. interValidity(spaceValidity(s1), timeValidity(t2)) = timespaceValidity(t2, s1)
. interValidity(spaceValidity(s1), timespaceValidity(t2, s2)) = timespaceValidity(t2, inter(s1, s2))
. interValidity(timespaceValidity(t1, s1), spaceValidity(s2)) = timespaceValidity(t1, inter(s1, s2))
. interValidity(timespaceValidity(t1, s1), timeValidity(t2)) = timespaceValidity(inter(t1, t2), s1)
. interValidity(timespaceValidity(t1, s1), timespaceValidity(t2, s2)) = 
         timespaceValidity(inter(t1, t2) , inter(s1, s2)) 
. testIntersectValidity(c1, c2) <=> not equal(interValidity(c1, c2), emptyValidity) 
%% equality axioms 

. incl(c1, c2) <=> incl(extractTime(c1), extractTime(c2)) 
 /\ incl(extractSpace(c1), extractSpace(c2)) 
 
%% theorems for setTime, setSpace, extractTime and extractSpace

. extractTime(setTime(c, t1)) = t1
. extractSpace(setSpace(c, s)) = s

. extractTime(interValidity(c1, c2)) = inter(extractTime(c1), extractTime(c2))
. extractSpace(interValidity(c1, c2))= inter(extractSpace(c1), extractSpace(c2))

. extractTime(union(c1, c2)) = union(extractTime(c1), extractTime(c2))
. extractSpace(union(c1, c2)) = union(extractSpace(c1), extractSpace(c2))

%% inclusion theorems
. incl(c1, c2) /\ incl(c2, c1)=> equal(c1, c2)
. incl(c1, c2) /\ incl(c2, c3) => incl(c1, c3)

%% equality theorems
. equal(c1, c1)
. equal(c1, c2) /\ equal(c2, c3) => equal(c1, c3)
. equal(c1, c2) => equal(c2, c1) 
. equal(c1, c2) <=> equal(extractTime(c1), extractTime(c2)) 
 /\ equal(extractSpace(c1), extractSpace(c2)) 

end




spec validityInstantTime =

Value

then

sort instantTime
sort duration 

%% generators
op undefined : instantTime 
op undefinedDuration: duration 
op instant : datavalue -> instantTime

op min : instantTime * instantTime -> instantTime
op max : instantTime * instantTime -> instantTime

pred equal : instantTime * instantTime
 
op union : instantTime * instantTime -> instantTime
op inter : instantTime * instantTime -> instantTime

pred testIntersect : instantTime * instantTime

pred __<__ : instantTime * instantTime 
pred __<=__ : instantTime * instantTime 

op __-__ : instantTime * instantTime -> duration 
op __+__ : instantTime * duration -> instantTime 

forall x, y : datavalue , i, i1, i2 : instantTime 
. min(instant(x), instant(y)) = instant(min(x,y))
. max(instant(x), instant(y)) = instant(max(x,y))

. min(i, i)=i
. max(i, i)=i
. min(instant(x), undefined) = undefined
. max(instant(x), undefined) = undefined
. min(undefined, instant(y)) = undefined
. max(undefined, instant(y)) = undefined

. instant(x) < instant(y) <=> min(x,y) = x
. instant(x) <= instant(y) <=> instant(x) < instant(y) \/ x = y
. equal(instant(x), instant(y)) <=> x = y

. i1 = i2 => union(i1, i2) = i1
. not(i1 = i2) => union(i1, i2) = undefined
. i1 = i2 => inter(i1, i2) = i1
. not(i1 = i2) => inter(i1, i2) = undefined
. i1 = i2 <=> testIntersect(i1, i2)

%%forall n, m : datavalue
%%.equal(instant(n), instant(m)) <=> (n = m)
%%.equal(undefined, undefined)

%% .instant(n) < instant(m) <=> n<m

end


spec ValidityTimeInterval =

validityInstantTime

then

sort timeInterval

%% Generators
op undefined : timeInterval   %% and undefined validity interval means valid at every time
op interval : instantTime * instantTime -> timeInterval
op interval : instantTime * duration -> timeInterval

pred equal : timeInterval * timeInterval 
pred disjoint : timeInterval * timeInterval
pred inside : instantTime * timeInterval
pred testIntersect : timeInterval * timeInterval
pred incl :  timeInterval * timeInterval


op union : timeInterval * timeInterval -> timeInterval
op interInterval : timeInterval * timeInterval -> timeInterval 

op startTime : timeInterval -> instantTime
op endTime : timeInterval -> instantTime
op duration : timeInterval -> duration

forall t1, t2: timeInterval, x, x1, x2, y1, y2: instantTime, d: duration
 
. equal(t1, t2) <=> equal(startTime(t1), startTime(t2)) /\ equal(endTime(t1), endTime(t2))

%% an undefined start time means -infinity, an undefined end time means +infinity
. inside(x, interval(x1, x2)) <=> not(x < x1) /\ not(x2 < x) 
. inside(x, interval(undefined, x2)) <=> not(x2 < x)
. inside(x, interval(x1, undefined)) <=> not(x < x1) 
. inside(x, interval(undefined, undefined))

. inside(x, interval(x1, d)) <=> not(x < x1) /\ not(x1+d < x)

. disjoint(t1 , t2) <=> not((startTime(t1) <= endTime(t2) /\ startTime(t2) <= endTime(t1)) 
    \/ (startTime(t2) <= endTime(t1) /\ startTime(t1) <= endTime(t2)))

. union(t1, undefined)=undefined
. union(undefined, t2)=undefined
. (not disjoint(interval(x1, x2), interval(y1, y2))) => 
 union(interval(x1, x2), interval(y1, y2)) = interval(min(x1, y1), max(x2, y2))

. startTime(interval(x1, x2)) = x1
. startTime(interval(x1, d)) = x1
. startTime(undefined) = undefined 

. endTime(interval(x1, x2)) = x2
. endTime(undefined) = undefined

. endTime(interval(x1, d)) = x1 + d
. endTime(interval(undefined, x2)) = undefined 
. endTime(interval(x1, undefined)) = undefined

. duration(interval(x1, x2)) = x2 - x1
. duration(interval(x1, d)) = d
. duration(undefined) = undefinedDuration

. incl(interval(x1, x2), t2) <=> inside(x1, t2) /\ inside(x2, t2)
. testIntersect(interval(x1, x2), t2) <=> inside(x1, t2) \/ inside(x2, t2)

. disjoint(t1, t2) => union(t1, t2) = undefined 

end


spec ValiditySpace =

sort geographicalRegion
sort country
sort city
sort state

%% generators
op geographicalRegion : country -> geographicalRegion
op geographicalRegion : city -> geographicalRegion
op geographicalRegion : state -> geographicalRegion

pred inside : geographicalRegion * geographicalRegion
pred testIntersectSpace : geographicalRegion * geographicalRegion
pred equal : geographicalRegion * geographicalRegion

op interSpace : geographicalRegion * geographicalRegion -> geographicalRegion
op union : geographicalRegion * geographicalRegion -> geographicalRegion


forall s1, s2, s3 : geographicalRegion
. inside(s1, s2) /\ inside(s2, s3) => inside(s1, s3) 
. inside(s1, s1)

. inside(s1, interSpace(s2, s3)) <=> (inside(s1, s2) /\ inside(s1, s3)) 
. inside(s1, union(s2, s3)) <=> (inside(s1, s2) \/ inside(s1, s3))

end

spec SequenceNode = NAT then
 Value
 then

sort sequenceNode 

%% generators
op emptySequence : sequenceNode
op seq : entity * entity -> sequenceNode 
op seq : entity * entity * nat -> sequenceNode 
op seqWithNext : entity -> sequenceNode
op seqWithPrev : entity -> sequenceNode 
op seqWithOrdinal : nat -> sequenceNode

%% operations
op next : sequenceNode -> entity
op previous : sequenceNode -> entity
op ordinal : sequenceNode -> nat

pred hasNext : sequenceNode 
pred hasPrevious : sequenceNode 
pred hasOrdinal : sequenceNode

forall x, y : entity, s : sequenceNode, n : nat
. next(seq(x, y)) = y
. previous(seq(x, y)) = x
. next(seqWithNext(x)) = x
. next(seqWithPrev(x)) = undefined
. previous(seqWithNext(x)) = undefined
. previous(seqWithPrev(x)) = x
. ordinal(seq(x, y, n)) = n
. ordinal(seqWithOrdinal(n)) = n
%%. hasPrevious(s) /\ ordinal(s) = n => ordinal(previous(s)) = n-1
%%. hasNext(s) /\ ordinal(s) = n => ordinal(next(s)) = n+1
 
. hasNext(seqWithNext(x)) 
. hasNext(seq(x, y)) 
. hasNext(seq(x, y, n)) 
. hasPrevious(seqWithPrev(x)) 
. hasPrevious(seq(x, y)) 
. hasPrevious(seq(x, y, n))  
. hasOrdinal(seq(x, y, n))
. hasOrdinal(seqWithOrdinal(n))
 

end



spec SET[sort Elem] =
  sort set[Elem]
  %%generator
  op emptyset : set[Elem]

  pred element : Elem * set[Elem]
  pred equal : set[Elem] * set[Elem]

  op __ union __ : set[Elem] * set[Elem] -> set[Elem]
  op add : Elem * set[Elem] -> set[Elem]
  op{__} : Elem -> set[Elem];

  forall x: Elem . {x} union {x} = {x}

end


spec Causality =

Value

then
SET[sort entity]
then
sort causality

%% generators
op emptyCause : causality
op addEndCause : set[entity] * causality -> causality 
op addHasCause : set[entity] * causality -> causality 

pred equal : causality * causality 

op getEndCause : causality -> set[entity]
op getHasCause : causality -> set[entity]
op unionCause : causality * causality -> causality
op inverseCause : causality -> causality
op inverseCause : entity -> entity

forall e1 : entity, c1, c2, c3 : causality

. equal(c1, c1) <=> equal(getEndCause(c1), getEndCause(c2)) 
                    /\ equal(getHasCause(c1), getHasCause(c2))

. getEndCause(emptyCause) = emptyset 
. getEndCause(addEndCause({e1}, c1))= add(e1, getEndCause(c1))
. getEndCause(addHasCause({e1}, c1))= getEndCause(c1)
. getEndCause(unionCause(c1, c2)) = getEndCause(c1) union getEndCause(c2)
. getEndCause(inverseCause(addEndCause({e1}, c1))) = add(inverseCause(e1), getEndCause(inverseCause(c1)))
. getEndCause(inverseCause(addHasCause({e1}, c1))) = getEndCause(inverseCause(c1))

. getHasCause(emptyCause) = emptyset
. getHasCause(addEndCause({e1}, c1))= getHasCause(c1)
. getHasCause(addHasCause({e1}, c1))= add(e1, getHasCause(c1))
. getHasCause(unionCause(c1, c2)) = getHasCause(c1) union getHasCause(c2)
. getHasCause(inverseCause(addEndCause({e1}, c1))) = getHasCause(inverseCause(c1))
. getHasCause(inverseCause(addHasCause({e1}, c1))) = add(inverseCause(e1), getHasCause(inverseCause(c1)))

%% axioms on inverseCause : entity -> entity are of the form
%% inverseCause(entity(wd:Q99521170)) = entity(wd:Q24037741) %% death of subject <--> death of subject's spouse
%% inverseCause(entity(wd:Q93190)) = entity(wd:Q93190) %% divorce <--> divorce

end


spec Provenance = Value then SET[sort entity] then

sort provenance

%% generators
op emptyProvenance : provenance
op addSources : set[entity] * provenance -> provenance

op getSources : provenance -> set[entity]
op union : provenance * provenance -> provenance

forall p1, p2 : provenance
. getSources(union(p1, p2)) = getSources(p1) union getSources(p2)


end


spec A1 = 
 sort s1
end

spec An =
 sort sn
end

spec Annotations = A1 and An

then
  SET[sort s1] and SET[sort sn]
then
sort annotations



%% Let A1, ..., An be the qualifiers that we consider as annotations
%% Let s1, ..., sn be the sorts of these qualifiers 
%% (it can be a datatype like string, int, real, ... or the sort resource)
%% we consider that each attribute may be multi-valued

%%generators
op emptyAnnotations : annotations 
op addA1 : annotations * s1 -> annotations
%% . . .
op addAn : annotations * sn -> annotations 

op getA1 : annotations -> set[s1] 
%% . . .
op getAn : annotations -> set[sn] 

forall a : annotations, v1 : s1 %{... }% , vn : sn

. getA1(addA1(a, v1)) = {v1} union getA1(a)
. getA1(emptyAnnotations) = emptyset

%% . . .

. getAn(addAn(a, vn)) = {vn} union getAn(a)
. getAn(emptyAnnotations) = emptyset



end


spec Validity = ValidityContext
     [ValidityTimeInterval 
           fit time |-> timeInterval,  inter |-> interInterval]
      [ValiditySpace  
           fit space |-> geographicalRegion, inter |-> interSpace] 

end

   
\end{lstlisting}
 
\end{appendices}

\end{document}